# Algorithms for Image Analysis and Combination of Pattern Classifiers with Application to Medical Diagnosis

Harris Georgiou[*]

National and Kapodistrian University of Athens
Department of Informatics and Telecommunications
`xgeorgio@di.uoa.gr`

**Abstract.** Medical Informatics and the application of modern signal processing in the assistance of the diagnostic process in medical imaging is one of the more recent and active research areas today. This thesis addresses a variety of issues related to the general problem of medical image analysis, specifically in mammography, and presents a series of algorithms and design approaches for all the intermediate levels of a modern system for computer-aided diagnosis (CAD). The diagnostic problem is analyzed with a systematic approach, first defining the imaging characteristics and features that are relevant to probable pathology in mammograms. Next, these features are quantified and fused into new, integrated radiological systems that exhibit embedded digital signal processing, in order to improve the final result and minimize the radiological dose for the patient. In a higher level, special algorithms are designed for detecting and encoding these clinically interesting imaging features, in order to be used as input to advanced pattern classifiers and machine learning models. Finally, these approaches are extended in multi-classifier models under the scope of Game Theory and optimum collective decision, in order to produce efficient solutions for combining classifiers with minimum computational costs for advanced diagnostic systems. The material covered in this thesis is related to a total of 18 published papers, 6 in scientific journals and 12 in international conferences.

**Keywords:** computer-aided diagnosis (CAD), medical image, mammography, pattern recognition, classifier combination, game theory.

## 1  Introduction

Advances in modern technologies and computers have enabled digital image processing to become a vital tool in conventional clinical practice, including mammography. However, the core problem of the clinical evaluation of mammographic tumors remains a highly demanding cognitive task. In order for these automated diagnostic systems to perform in levels of sensitivity and specificity similar to that of human experts, it is essential that a robust framework on problem-specific design parameters is formulated.

---

[*] Dissertation Advisor: Sergios Theodoridis, Professor.

## 2 Qualitative Features for Mammographic Diagnosis

### 2.1 Overview of Work

Advances in modern technologies and computers have enabled digital image processing to become a vital tool in conventional clinical practice, including mammography. However, the core problem of the clinical evaluation of mammographic tumors remains a highly demanding cognitive task. In order for these automated diagnostic systems to perform in levels of sensitivity and specificity similar to that of human experts, it is essential that a robust framework on problem-specific design parameters is formulated.

This study is focused on identifying a robust set of clinical features that can be used as the base for designing the input of any computer-aided diagnosis system for automatic mammographic tumor evaluation. A thorough list of clinical features was constructed and the diagnostic value of each feature was verified against current clinical practices by an expert physician. These features were directly or indirectly related to the overall morphological properties of the mammographic tumor or the texture of the fine-scale tissue structures as they appear in the digitized image, while others contained external clinical data of outmost importance, like the patient's age. The entire feature set was used as an annotation list for describing the clinical properties of mammographic tumor cases in a quantitative way, such that subsequent objective analyses were possible.

For the purposes of this study, a mammographic image database was created, with complete clinical evaluation descriptions and positive histological verification for each case. All tumors contained in the database were characterized according to the identified clinical features' set and the resulting dataset was used as input for discrimination and diagnostic value analysis for each one of these features. Specifically, several standard methodologies of statistical significance analysis were employed to create feature rankings according to their discriminating power. Moreover, three different classification models, namely linear classifiers, neural networks and support vector machines, were employed to investigate the true efficiency of each one of them, as well as the overall complexity of the diagnostic task of mammographic tumor characterization.

## 2.2 Results and Discussion

The study was based on four distinct issues: (1) Create a thorough list of abnormal findings regarding diagnostic evaluation of a mammogram, especially related to image textural and morphological features of the underlying tissue. From this list, the most prominent and content-rich features were to be selected, according to their suitability for automatic extraction through image processing algorithms. (2) Create a specialized mammographic image database, containing clearly identifiable and histologically verified cases of benign or malignant tumors. All cases were evaluated and annotated in relation to the previously defined list of important clinical features. (3) Analyze the newly constructed set of mammographic images in relation to the feature list, focusing especially on investigating the importance, comprehensiveness and consistency of each one of these features when correlated with the verified final diagnosis. (4) Investigate the performance of individual features, as well as subsets of combined features, when used as real training datasets for various classifier architectures, i.e., linear classifiers, neural networks (NN) and support vector machines (SVM).

The final set of 9 clinical features was the base for the annotation list, which was used to describe and document the expert's clinical evaluation for each mammographic image in the database. Specifically, (1) the *presence of tumors*, (2) the presence of *microcalcifications*, (3) the *tumor density*, (4) the *percentage of fat* within the tumor, (5) the *tumor boundary vagueness*, (6) the *tumor homogeneity*, (7) the tumor *morphological shape type*, (8) the *patient's age*, as well as (9) the final *histological diagnosis*, were included. As the patient's age remains a feature of high clinical importance, it was also included in the final annotation list as a unique "external" data, although it cannot be referred directly from the mammographic image itself. Finally, the morphological shape type refers to the classification of the tumor's shape in one of four predefined shape categories, related to tumor's boundary roughness and stellate or lobulated outline.

Both the statistical and the classification results have proven the explicit correlation of all the selected features with the final diagnosis, qualifying them as an adequate input base for any type of similar automated diagnosis system. The underlying complexity of the diagnostic task has justified the high value of sophisticated pattern recognition architectures.

Preliminary analysis on the initial dataset has confirmed the strong statistical correlation between morphological shape type and verified diagnosis of breast tumors in the mammograms. Specifically, the first two types of morphology, round and lobulated tumors, exhibited 17% and 5% of malignancy, respectively, within the same class. On the contrary, micro-lobulated and stellate types exhibited 95% and 97% of malignancy, respectively. When combining the round and lobulated cases, the overall percentage of malignancy was 12%, while for combined micro-lobulated and stellate cases, the overall percentage of malignancy was 96%. This high statistical dependency of specific morphological features of each tumor with its verified pathology confirms the clinical value of its shape when conducting a pathological evaluation of a mammogram. It should be noted that if the shape type feature were to be used as the sole input for predicting the final diagnosis, an accuracy rate just over 93% could be achieved.

Statistical and classification analysis results have shown that, although the selected feature sets were in fact content-rich with regard to their diagnostic value, the diagnostic process itself remains a complex and demanding task. The high degree of non-linearity employed in the discrimination of the input data with regard to diagnosis prediction suggests that automatic diagnosis systems should implement powerful pattern recognition models of non-linear and highly adaptive architecture. Future work should be focused on designing specialized image processing algorithms for efficient automatic extraction of morphological and textural features, combined with robust implementations of advanced classification architectures, such as SVMs. Automated diagnosis of breast mammographic abnormalities, combined with CAD systems, which indicate suspicious lesions in mammograms, will be a very powerful tool in the hands of the mammographic departments and the reporting physicians, especially the less experienced ones.

## 3  Content-Based Automatic Exposure Rate Control in Mammography

### 3.1  Overview of Work

I-ImaS (Intelligent Imaging Sensors) is an EU FP6 funded project aiming to produce new, intelligent, specialised X-ray imaging systems utilising novel Monolithic Active Pixel Sensors (MAPS) to create optimal diagnostic images. MAPS technology allows on-chip processing; leading to the potential to provide on-line analysis of data during image acquisition that can be used as an input into an on-line exposure modulation feedback control.

Current diagnostic X-ray systems allow the selection of global exposure settings such as beam energy and flux. Digital images may be acquired using either an area detector or a linear detector; in the latter case, a scan is implemented moving the detector under the sample. In a recent paper discussing the development of AEC (Automatic Exposure Control) in scan systems, the idea of a modulation of the scan velocity to optimize the signal in each image row was introduced.

In our system a linear detector is used and the exposure is optimized in each image sub-region (about 1 x 16 mm$^2$): the beam intensity is a function not only of tissue thickness and attenuation, but also of local physical and statistical parameters found in the image itself. Using a linear array of detectors with on-chip intelligence, we are developing a system that performs on-line analysis of the image and optimizes the X-ray intensity during the scanning procedure to obtain the maximum diagnostic information from the region of interest while minimizing exposure in the regions where a good image quality is obtained with low fluence.

## 3.2 Results and Discussion

The preliminary image analysis was based on publicly available, well-documented sets of mammographic images. A complete and thorough documentation set of templates was designed in order to record and file all equipment-related information during image acquisition experiments, as well as the technical and clinical aspects of image quality assessments.

A simulation model was created for processing initial mammographic images, acquired at optimal exposure settings, and produce simulated versions of the same image at overexposed and underexposed conditions. The model was verified by comparing its results with a real independent set of phantom images, acquired at similar overexposure and underexposure settings. Simulated image sets, along with the initial (optimal) images, were used as the base for a texture-analysis model. This new model was used to implement a full set of 20 textural feature functions, including 1st order statistics, signal "roughness" metrics, as well as a set of specially designed synthetic features. Image processing and feature values extraction was adapted according to the real functionality and data acquisition of the final line-scanning system.

Intermediate 2-D texture matrices from texture-analysis model were reduced to 1-D simple curves and combined with the simulation model for acquiring all texture features at different exposure settings. The combined system produced clear results on the main issue of establishing a consistent functional link between texture analysis and exposure conditions. A set of 8 most prominent candidates, from the total of 20 textural feature functions, was identified as well-suited for this task. Finally, the preliminary textural feature results and selections were confirmed by applying the same analytical procedure over a new image database, containing X-ray images of real breast tissue samples, exposed at various ranges of kVp and mAs. Implementation of all feature functions was confirmed to be fully compatible with the sensor IC requirement for SIMD h/w architecture. However, before the choices for feedback control are finalized, further verification and adjustment test have to conducted, using real (instead of simulated) sets of sub-optimally exposed X-ray images.

The control system is based on the I-Imas system architecture and specifications. In this architecture, the tissue is first imaged using a so-called "scout scan". This initial imaging is done using a constant dose. The data from the scout scan is analyzed automatically in real-time in order to determine the correct exposure for the second scan. In order to develop a control system the images in the developed database were evaluated by 2 or 3 radiologists. An analysis of different options for embedding different levels of "intelligence" into the sensor IC, from basic linear adaptive control to sophisticated classifier-based non-linear modules has been performed. This context has been used to select a good framework for developing the control algorithm.

The study includes prototyping of an initial control system algorithm. The algorithm is based on data from a scout scan combined with a pre-established sensor model to predict the correct exposure. The main goal of this algorithm is to avoid underexposure. The algorithm goal was chosen amongst many other candidates, and give consistently good results on the available images. The regulated images from the mammography database were evaluated together with a reference image in a blind test by two experienced radiologists at Ulleval University Hospital in Norway. The reference image was taken with uniform exposure at the highest dose used in the regulated image. All the regulated images were evaluated to have the same or better overall quality than the reference image, however at a lower dose.

The development and the evaluation of the steering algorithm are based on custom tissue images, which consist of a limited number of tissue samples. Consequently, there is some uncertainty considering the robustness of the algorithm. There has also been developed a visualization system, which makes the images from the developed sensor appear as though they were captured with a normal, constant-exposure sensor system. This visualization model is based on a model of the sensor.

The results from the prototype system show that the detectors to be used in the I-ImaS system are capable of producing good diagnostic quality images. The prototype system, which operates a full 10 sensor long detector array with intelligence driven step-wedge modulation of the x-ray beam quality has now been constructed based on this pre-prototype system and is currently undergoing acceptance testing and commissioning. It is hoped that this prototype system will prove to be capable of producing intelligent images containing an increased level of diagnostic information compared to current AEC systems.

## 4 Mammographic Masses Characterization based on Localized Texture

### 4.1 Overview of Work

Localized texture analysis of breast tissue on mammograms is an issue of major importance in mass characterization. However, in contrast to other mammographic diagnostic approaches, it has not been investigated in depth, due to its inherent difficulty and fuzziness. This work aims to the establishment of a quantitative approach of mammographic masses texture classification, based on advanced classifier architectures and supported by fractal analysis of the dataset of the extracted textural features. Additionally, a comparison of the information content of the proposed feature set with that of the qualitative characteristics used in clinical practice by expert radiologists is presented.

An extensive set of textural feature functions was applied to a set of 130 digitized mammograms, in multiple configurations and scales, constructing compact datasets of textural "signatures" for benign and malignant cases of tumors. These quantitative textural datasets were subsequently studied against a set of a thorough and compact list of qualitative texture descriptions of breast mass tissue, normally considered under a typical clinical assessment, in order to investigate the discriminating value and the statistical correlation between the two sets. Fractal analysis was employed to compare the information content and dimensionality of the textural features datasets with the qualitative information provided through medical diagnosis. A wide range of linear and non-linear classification architectures was employed, including linear discriminant analysis (LDA), least-squares minimum distance (LSMD), K-nearest-neighbors (K-nn), radial basis function (RBF) and multi-layer perceptron (MLP) artificial neural network (ANN), as well as support vector machine (SVM) classifiers. The classification process was used as the means to evaluate the inherent quality and informational content of each of the datasets, as well as the objective performance of each of the classifiers themselves in real classification of mammographic breast tumors against verified diagnosis.

**4.2 Results and Discussion**

Textural features extracted at larger scales and sampling box sizes proved to be more content-rich than their equivalents at smaller scales and sizes. Fractal analysis on the dimensionality of the textural datasets verified that reduced subsets of optimal feature combinations can describe the original feature space adequately for classification purposes and at least the same detail and quality as the list of qualitative texture descriptions provided by a human expert. Non-linear classifiers, especially SVMs, have been proven superior to any linear equivalent. Breast mass classification of mammograms, based only on textural features, achieved an optimal score of 83.9%, through SVM classifiers.

The two texture datasets for 20-pixel and 50-pixel box sizes were used separately in comparative training configurations. For the 20-pixel box dataset, the results of classification accuracy ranged from 62.6% to 80.4% according to the exact classifier selection. Both the LDA and LSMD classifiers achieved only the lowest performance of 62.6% even when using optimized combinations of features. Next, the MLP and RBF neural classifiers with optimized topology scored 74.4% and 71.3% correspondingly. The overall best accuracy was achieved by the SVM classifier at 80.4%, followed very closely by the optimized K-nn classifier at 80.3%. For the 50-pixel dataset, the results of classification accuracy ranged from 69.0% to 83.9% according to the exact classifier selection. As in the first dataset, both LDA and LSMD achieved the lowest score equally at 69.0%. The RBF neural classifier achieved 72.8% and the MLP classifier outperformed it with better accuracy at 78.2%. Again, the overall best performance was achieved by the SVM classifier at 83.9%, followed closely by the optimized K-nn classifier at 83.6%.

Classification results over linear models, employing exhaustive feature combination optimization, have provided some indications regarding the most appropriate textural feature functions and their configurations for local processing, for classification problems against the verified diagnosis of the complete tumor. As the problem becomes too complex for simple linear systems, more efficient structures are necessary in order to exploit the complete range of the discriminating power of the available texture datasets. MLP neural classifiers outperformed all other linear and ANN architectures, while K-nn and especially SVM classifiers achieved the overall best accuracy rates.

## 5 Multi-Scaled Morphological Features for Mammographic Masses

### 5.1 Overview of Work

A comprehensive signal analysis approach on the mammographic mass boundary morphology is presented in this article. The purpose of this study is to identify efficient sets of simple yet effective shape features, employed in the original and multi-scaled spectral representations of the boundary, for the characterization of the mammographic mass. These new methods of mass boundary representation and processing in more than one domain greatly improve the information content of the base data that is used for pattern classification purposes, introducing comprehensive spectral and multi-scale wavelet versions of the original boundary signals. The evaluation is conducted against morphological and diagnostic characterization of the mass, using statistical methods, fractal dimension analysis and a wide range of classifier architectures.

This study consists of: (a) the investigation of the original radial distance measurements under the complete spectrum of signal analysis, (b) the application of curve feature extractors of morphological characteristics and the evaluation of the discriminative power of each one of them, by means of statistical significance analysis and dataset fractal dimension, and (c) the application of a wide range of classifier architectures on these morphological datasets, in order to conduct a comparative evaluation of the efficiency and effectiveness of all architectures, for mammographic mass characterization. Radial distance signal was exploited using the discrete Fourier transform (DFT) and the discrete wavelet transform (DWT) as additional carrier signals. Seven uniresolution feature functions were applied over these carrier signals and multiple shape descriptors were created. Classification was conducted against mass shape type and clinical diagnosis, using a wide range of linear and non-linear classifiers, including linear discriminant analysis (LDA), least-squares minimum distance (LSMD), k nearest neighbor (k-NN), radial basis function (RBF) and multi-layered perceptron (MLP) neural networks (NN), and support vector machines (SVM). Fractal analysis was employed as a dataset analysis tool in the feature selection phase. The discriminative power of the features produced by this composite analysis is subsequently analyzed by means of multivariate analysis of variance (MANOVA) and tested against two distinct classification targets, namely (a) the morphological shape type of the mass and (b) the histologically verified clinical diagnosis for each mammogram.

## 5.2 Results and Discussion

Statistical analysis and classification results have shown that the discrimination value of the features extracted from the DWT components and especially the DFT spectrum, are of great importance. Furthermore, much of the information content of the curve features in the case of DFT and DWT datasets is directly related to the texture and fine-scale details of the corresponding envelope signal of the spectral components. Neural classifiers outperformed all other methods (SVM not used because they are mainly two-class classifiers) with overall success rate of 72.3% for shape type identification, while SVM achieved the overall highest 91.54% for clinical diagnosis. Receiver operating characteristic (ROC) analysis has been employed to present the sensitivity and specificity of the results of this study.

Comparative results over all the available datasets and feature groupings have shown that the effectiveness of each feature function depends greatly on each specific base signal under investigation, as well as the intended classification target, i.e., mass shape type or diagnosis. Among the total of 217 features collected from all the carrier signals, i.e., radial distance measurements, DFT spectrum envelope and DWT components, MANOVA analysis was applied to select the ones with maximum independence and outmost statistical significance. For these optimal feature selections, both for the shape type and the clinical diagnosis target cases, specific curve feature functions were favored at some degree, like zero-cross count (for shape type only), roughness index (for diagnosis only), mean value and area ratio (for both classification targets). On the contrary, some feature functions like circularity seems to be of little or no use in terms of discriminative power. However, combined results over all the optimal classifier configurations showed no strong evidence supporting or rejecting the selection of any specific curve feature function, either for any specific carrier signal or for hybrid signal configurations.

Using various feature sub-sets with optimized linear and neural classifier models, it has been established that the DFT spectrum and the DWT components capture discriminative information of significant importance, in relation to both morphological shape type and clinical diagnosis. Furthermore, NN and SVM classifiers outperformed linear ones in all cases, while least-squares based classifiers exhibited the highest accuracies over the linear architectures. The use of spectral properties and wavelet components of the original radial distance signal, in conjunction with optimized NN and SVM classifiers, produced significant increase in the overall success rate of the diagnostic system.

# 6 A Game-Theoretic Approach for Combining Pattern Classifiers

## 6.1 Overview of Work

A new approach from the game-theoretic point of view is proposed for the problem of optimally combining classifiers in dichotomous choice situations. The analysis of weighted majority voting under the viewpoint of coalition gaming, leads to the existence of analytical solutions to optimal weights for the classifiers based on their prior competencies.

The game-theoretic modeling of combining classifiers in dichotomous choice problems leads to cooperative gaming approaches, specifically coalition gaming in the form of weighted majority rules (WMG). Theoretically optimal solutions for this type of games are the WMR schemes, often referred to as weighted majority voting. Under the conditional independence assumption for the experts, there exists a closed solution for the optimal weighting profiles for the WMR formula.

The general framework of WMR is tested against common rank-based and simple majority models, as well as two soft-output averaging rules. Experimental results with combined support vector machine (SVM) classifiers on benchmark classification tasks have proven that WMR, employing the theoretically optimal solution for combination weights proposed in this work, outperformed all the other rank-based, simple majority and soft-output averaging methods. It also provides a very generic and theoretically well-defined framework for all hard-output (voting) combination schemes between any type of classifier architecture.

## 6.2 Results and Discussion

The evaluation of the combination models consisted of two phases, namely: (a) the design and training of SVM classifiers, trained in distinctly different subspaces, and (b) the application of the various combination schemes to the outputs of the individual classifiers. Each of the K classifiers was separately trained and optimized, using a different group of features from the full dataset, and subsequently evaluated using the corresponding validation set. This training/validation cycle was applied three times, for each of the five datasets, each time using a new random partitioning of the full dataset into training and validation sets.

Experimental comparative results have shown that such simple combination models for ensembles of classifiers can be more efficient than all typical rank-based and simple majority schemes, as well as simple soft-output averaging schemes in some cases. Specifically, the all versions of the WMR model exhibited the best performance amongst all the other hard-output combination rules. As expected, it has been proven better than the simple majority voting, as well as all the other rank-based methods (max, min, median). The "odds" weighting profile has also been proven marginally better than the "direct"- and the "logodds"-based profiles for designing the optimal WMR formula.

Although the conditional independence assumption was moderately satisfied by using distinct partitions of the feature space, results have shown that the theoretical solution is still valid to a considerable extent. Therefore, the WMR can be asserted as a simple yet effective option for combining almost any type of classifier with others in an optimal and theoretically well-defined framework.

# 7 Conclusions

The current PhD thesis presents a wide variety of issues related to the general problem of medical image analysis, specifically in digital mammography.

Firstly, the qualitative clinical features for mammographic diagnosis were identified and analyzed by statistical and pattern classification methods. Next, new methodologies were investigated on the used of low-level imaging data for content-based automatic exposure control in mammographic systems. Higher-level imaging analysis of mammographic tumors for automated diagnosis was investigated under the scope of region-based (texture) and contour-based (shape) classification, using a wide variety of statistical, fractal dimensionality and pattern classification methods. Finally, the classification phase was extended within the scope of combining pattern classifiers, introducing a novel game-theoretic approach as the base aggregation scheme and evaluating it comparatively against other models, with promising results.

Future work should be focused on designing specialized image processing algorithms for more efficient automatic extraction of morphological and textural features, combined with robust implementations of advanced classification architectures, such as ensembles of pattern classifiers. Automated diagnosis of breast mammographic abnormalities, combined with CAD systems, which indicate suspicious lesions in mammograms, will be a very powerful tool in the hands of the mammographic departments and the reporting physicians, especially the less experienced ones.